\documentclass{article}

\usepackage{natbib}
\usepackage{dsfont}
\usepackage[ruled, longend, linesnumbered]{algorithm2e}
\usepackage{amsmath}
\usepackage{amssymb}
\usepackage[pdftex]{graphicx}
\usepackage[colorlinks, citecolor=blue]{hyperref}
\usepackage{tikz-qtree}
\usepackage{multirow}
\usepackage{multicol}
\usepackage{colortbl}
\usepackage{arydshln}
\usepackage{mathtools}
\usepackage{ulem}
\usepackage{xcolor}

\newcommand{\bZ}{\mathbf{Z}}
\newcommand{\bz}{\mathbf{z}}
\newcommand{\cA}{\mathcal{A}}
\newcommand{\cT}{\mathcal{T}}
\newcommand{\bp}{\mathbf{p}}
\newcommand{\bs}{\mathbf{s}}
\newcommand{\cz}{\widetilde{\mathbf{z}}}
\newcommand{\cZ}{\widetilde{\mathbf{Z}}}
\newcommand{\balpha}{\boldsymbol{\alpha}}
\newcommand{\cE}{\mathcal{E}}
\newcommand{\train}{\text{train}}
\newcommand{\test}{\text{test}}
\newcommand{\niter}{{n_\text{iter}}}
\newcommand{\BF}{\text{BF}}
\newcommand{\AIBF}{\text{AIBF}}
\newcommand{\GIBF}{\text{GIBF}}

\newcommand{\grow}{{\oplus}}
\newcommand{\ungrow}{{\ominus}}

\newtheorem{definition}{Definition}

\title{Detecting Renewal States in Chains of Variable Length via Intrinsic Bayes Factors}
\date{}
\author{
  Victor Freguglia and  Nancy L. Garcia \\
  {\small Department of Statistics -- University of Campinas} }

\begin{document}
\maketitle

\begin{abstract}
Markov chains with variable length are useful parsimonious stochastic models able to generate most stationary sequence of discrete symbols.
The idea is to identify the suffixes of the past, called contexts, that are relevant to predict the future symbol.
Sometimes a single state is a context, and looking at the past and finding this specific state
makes the further past irrelevant. States with such property are called renewal states and they can be used to split the chain into independent and identically distributed blocks.
In order to identify renewal states for chains with variable length, we propose the use of Intrinsic Bayes Factor to evaluate the hypothesis that
some particular state is a renewal state.
In this case, the difficulty lies in integrating the marginal posterior distribution for the random context trees for general prior distribution on the space of context trees,
with Dirichlet prior for the transition probabilities, and Monte Carlo methods are applied.
To show the strength of our method, we analyzed artificial datasets generated from different binary models models and one example coming from the field of Linguistics.
\end{abstract}

\section{Introduction}

Markov Chains with variable length are useful stochastic models that provide a powerful framework
for describing transition probabilities for finite-valued sequences
due the possibility of capturing long-range interactions while keeping
some parsimony in the number of free parameters.
These models were introduced in the seminal paper of \citet{rissanen1983universal} for data compression
and became known in the statistics literature as Variable Length Markov Chain (VLMC) by  \citet{buhlmann1999variable},
and as Probabilistic Suffix Trees (PST) in the machine learning literature \citep{ron1996power}.
The idea is that, for each past, only a finite suffix of the past is enough to predict the next symbol.
Rissanen called {\it context}, the relevant ending string of the past.
The set of all contexts can be represented by the set of leaves of a rooted tree
if we require that no context is a proper suffix of another context.
For a fixed set of contexts, estimation of the transition probabilities can be easily achieved.
The problems lies into estimating the set of contexts from the available data.
In his seminal 1983 paper, Rissanen introduced the Context algorithm,
which estimates the context tree by aggregating irrelevant states in the history of the process using a sequential procedure.
A nice introductory guide to this type of models and particularly to the Context algorithm can be found in \citet{galves2008stochastic}.

Many of the tree model methods related to data compression tasks involve obtaining
better predictions based on weighting over multiple models.
A classical example is the Context-Tree Weighting (CTW) algorithm \citep{willems1995context},
which computes the marginal probability of a sequence by weighting over all context trees
and all probability vectors using computationally convenient weights.
Using CTW, \citet{csiszar2006context} showed that context trees can be consistently estimated
in linear time using the Bayesian information criterion (BIC).
These weighting strategies can be translated to a Bayesian framework
where unobserved parameters of a probabilistic system are treated as additional random components
with given prior distribution and inference is based on integrating over the nuisance parameters,
which is a form of weighting over these quantities based on the prior distribution.
Nonetheless, inference performed following the Bayesian paradigm
for VLMC models is a relatively recent topic of research.
Some works that explicitly use Bayesian statistics in combination with VLMC models are
\citet{dimitrakakis2010bayesian} which introduced an online prediction scheme by adding a prior,
conditioned on context, on the Markov order of the chain,
and \citet{kontoyiannis2020bayesian} which provided more general tools such as posterior
sampling through Metropolis-Hastings algorithm and Maximum a Posteriori context tree estimation
focusing on model selection, estimation, and sequential prediction.
A Bayesian approach for model selection in high-order Markov chains,
allowing conditional probabilities to be partitioned into more general structures
than the tree-based structures of VLMC models, is also proposed in \citet{xiong2016recursive}.

As aforementioned, the effort was mostly concentrated in estimating the context tree structure.
On the other hand, hypothesis testing for VLMC is a difficult topic, first tackled by \citet{balding2009limit}
and pursued further by \citet{busch2009testing} using a Kolmogorov-Smirnov-type goodness-of-fit test,
to compare if two samples come from the same distribution.
Under the Bayesian paradigm, hypothesis testing is done though Bayes Factor, but computing Bayes Factors
may depend on integrations that require enormous computational effort depending on the random objects
and hypotheses involved.
Particularly for VLMC problems, Bayes Factors require summing over the set of all possible context trees,
which cardinality grows doubly exponentially with the maximum depth considered, quickly becoming intractable.
To avoid such intractable quantities, we use the Monte Carlo approximations of the Intrinsic Bayes Factor
\citep{berger1996intrinsic}, which is based on averaging over posterior distributions,
that tend to be highly concentrated within a small set of context trees for VLMC models,
and have been used recently in many different fields with the same purpose,
such as \citet{cabras2015new}, \citet{charitidou2018objective} and \citet{villa2021objective}.
Alternatives to Bayes Factors based on using posterior distributions instead of the prior distribution
in integrations have been evolving over the past decades, the classical method using this strategy
is the Posterior Bayes Factor \citet{aitkin1991posterior}, with applications in a variety of models
such as \citet{aitkin1993posterior} and \citet{aitkin1996new}.

In this work, we focus on one characteristic of interest in a context tree, the presence of a renewal state or a renewal context.
Renewal states play an important role in some computational methods frequently used in statistical analysis
such as designing Bootstrap schemes and defining proper cross-validation strategies based on blocks.
Therefore, having some methodology not only to detect renewal states, but also to quantify how plausible these assumptions are,
can improve the robustness of analysis at the cost of some pre-processing computations.
For example, \citet{galves2012context} proposed a constant free algorithm (Smallest Maximizer Criterion) to find the tree that maximizes the BIC based on a Bootstrap scheme that uses the renewal property of one of the states.  To the best of our knowledge, using Bayes Factors for evaluating hypotheses involving probabilistic context trees is a topic that has not been explored.


\section{Variable-Length Markov Chains} \label{sec:vlmc}

\subsection{Model Description}

Let $\cA$ be an {\it alphabet} of $m$ symbols, and without loss of generality, consider $\cA = \{0, 1, \dots, (m-1)\}$ for simplicity.
For $t_2 > t_1$, a string  $(z_{t_1}, \ldots,z_{t_2}) \in \cA^{t_2-t_1+1}$ will be denoted by $\bz_{t_1}^{t_2}$ and its length by $\ell(\bz_{t_1}^{t_2}) = t_2-t_1+1$.
A sequence $\bs_{n-l}^{n}$ is a {\it suffix} of a string $\bz_{1}^{n}$ if $s_j = z_j$ for all $j=n-l, \ldots, n$. If $l < n$ we say that $s_{n-l}^{n}$ is a {\it proper suffix} of the string $\bz_{1}^{n}$.

\begin{definition}
  Let $L>0$ and $\tau \subset \cup_{j=1}^{L} \cA^{j}$ be a set of strings formed by symbols in $\cA$.
  We say that $\tau$ satisfies the {\rm suffix property} if, for
  every string  $\bs_{-j+j'}^{-1} = (s_{-j+j'}, \dots, s_{-1}) \in \cA^{j-j'}$, $\bs_{-j+j'}^{-1}\in\tau$ implies that $\bs_{-j}^{-1} \not\in \tau$ for $j > 1, j'=1,\dots,j$.
\end{definition}

\begin{definition}
  Let $L > 0$ and $\tau \subset \cup_{j=1}^{L} \cA^{j}$ be a set of strings formed by symbols in $\cA$.
  We say that $\tau$ is an {\rm  irreducible tree} if, no string belonging to $\tau$ can be
  replaced by a proper suffix without violating the suffix property.
\end{definition}

\begin{definition}
  Let $\tau$ be an irreducible tree. We say that $\tau$ is {\rm full} if, for each string $\bs \in \tau$, any concatenation of a symbol $k \in \cA$ and a suffix of $\bs$ is the suffix of a string $\bs' \in \tau$.
\end{definition}

\paragraph*{Examples}

Suppose that we have a binary alphabet $\cA = \{0,1\}$, then:
\begin{itemize}
  \item $\tau = \{0, 1, 11\}$ does \textbf{not satisfy the suffix property} because it contains both the strings $1$ and $11$.
  \item $\tau = \{0, 01\}$ is \textbf{not an irreducible tree}, because the string $01$ can be replace by its suffix, $1$, without violating the suffix property, as the set $\{0,1\}$ satisfies the suffix property.
  \item $\tau = \{0, 011, 111\}$ \textbf{is irreducible}, but it is \textbf{not full} because $1$ is a suffix of a string in $\tau$ (either $011$ or $111$), but $01$ (the concatenation of $0 \in \cA$ and $1$) is not.
  \item $\tau_1 = \{0, 100, 101, 110, 111\}$, $\tau_2 = \{01, 00, 10, 11\}$ and $\tau_3 = \{0, 10, 110, 1110, 1111\}$ are  \textbf{full irreducible} trees.
\end{itemize}

A full irreducible tree $\tau$ can be represented by the set of leaves of a rooted tree with a finite set of labeled
branches such that
\begin{enumerate}
\item [(1)] The root node has no label,
\item [(2)] each node has either $0$ or $m$ children (fullness) and
\item [(3)] when a node has $m$ children, each child has symbol of the alphabet $\cA$ as a label.
\end{enumerate}

\begin{figure}[!htbp]
  \centering
  \begin{tikzpicture}[grow=down, scale=0.66]
  \tikzset{edge from parent/.style={draw,edge from parent path={(\tikzparentnode.south)-- +(0,-8pt)-| (\tikzchildnode)}},
    every internal node/.style={draw,circle}}
    \Tree [ .{}
    [ .0 ]
        [ .1
            [ .0 ]
            [ .1
              [ .0 ]
              [ .1
                [ .0 ]
                [ .1
                  [ .0 ]
                  [ .1 ]]]]]]
                  \node[above] at (current bounding box.north) {(I)};
  \end{tikzpicture}
  \begin{tikzpicture}[grow=down, scale=0.66]
  \tikzset{edge from parent/.style={draw,edge from parent path={(\tikzparentnode.south)-- +(0,-8pt)-| (\tikzchildnode)}},
    every internal node/.style={draw,circle}}
    \Tree [ .{}
    [ .0 ]
        [ .1
        [ .0 [ .0 ] [ .1 ] ]
            [ .1
              [ .0 ]
              [ .1
                [ .0 [ .0 ] [ .1 ] ]
                [ .1
                  [ .0 ]
                  [ .1 ]]]]]]
                  \node[above] at (current bounding box.north) {(II)};
  \end{tikzpicture}
  \begin{tikzpicture}[grow=down, scale=0.66]
  \tikzset{edge from parent/.style={draw,edge from parent path={(\tikzparentnode.south)-- +(0,-8pt)-| (\tikzchildnode)}},
    every internal node/.style={draw,circle}}
    \Tree [ .{}
    [ .0 ]
    [ .1 ]
    [ .2 [ .0 [ .0 ]
              [ .1 ]
              [ .2 ]
              [ .3 ] ]
         [ .1 ]
         [ .2 ]
         [ .3 ]]
    [ .3 [ .0 ]
         [ .1 ]
         [ .2 [ .0 ]
              [ .1 ]
              [ .2 ]
              [ .3 ]]
         [ .3 ]]]
                  \node[above] at (current bounding box.north) {(III)};
  \end{tikzpicture}
  \caption{Examples of context trees.}\label{fig:renewal_examples}
\end{figure}
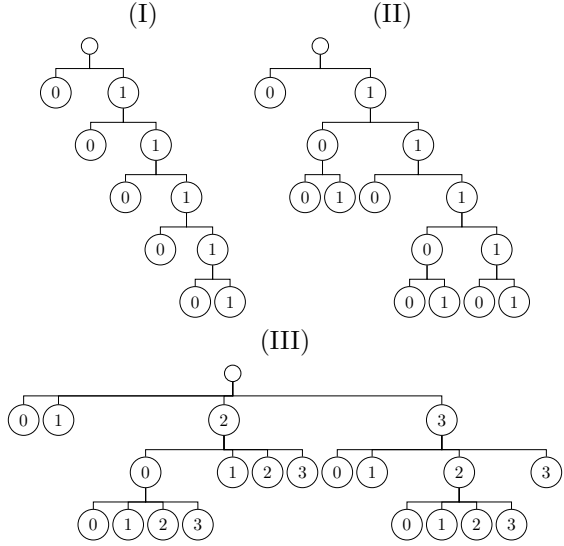

The elements of $\tau$ will be called {\it contexts} and we will refer to full irreducible trees as {\it context trees} henceforth. \autoref{fig:renewal_examples} presents 3 examples of contexts trees.  The {\it depth} $\ell$ of a tree $\tau$ is given by the maximal length of a context belonging to $\tau$, defined as
$$\ell(\tau) = \max\{\ell(\bz); \bz \in \tau\}.$$
In this work we will assume that the depth of the tree is  bounded by an integer $L$.  In this case, it is straightforward to conclude that,  for any string $\bz_{t_1}^{t_2}$
with at least $L+1$ symbols,
there exist a suffix $\bz_{t_2 - l}^{t_2}$ and a leaf of $\tau$ such that the symbols between the leaf
(including) and the root node are exactly $\bz_{t_2 - l}^{t_2}$.
\citet{galves2012context} referred to this property as the {\it properness} of a context tree.

For each context tree $\tau$, we can associate a family of probability measures indexed by elements of $\tau$,
$$\bp = \{ p(\cdot | \bs): \cA \rightarrow [0,1]; \bs \in \tau\}.$$
The pair $(\tau,\bp)$ is called a {\it probabilistic context tree}.

Given a tree $\tau$ with the described properties and depth bounded by $L$,
define a {\it suffix mapping function} $\eta_{\tau}: \cup_{j=L+1}^{\infty} \cA^{j} \rightarrow \tau$ such that $\eta_\tau(\bz_{t_1}^{t_2}) = \bz_{t_2-l}^{t_2}$ is the unique suffix
$\bz_{t_2-l}^{t_2} \in \tau$.

\begin{definition} A sequence of random variables  $\bZ = (Z_t)_{t = 1}^T$ with state space $\cA$ is a {\it Variable Length Markov Chain (VLMC)} compatible with the probabilistic tree
$(\tau,\bp)$ if it satisfies
\begin{equation}\label{eq:cond_prob}
  \mathds{P}\left(Z_t = k | \bZ_1^{t-1} = \bz_{1}^{t-1}\right) =
   \bp(k|\eta_\tau(\bz_1^{t-1})),
\end{equation}
for all $L < t \le T$, $ \bz_{1}^{t-1} \in \cA^{t-1}$, where $ \eta_\tau(\bz_1^{t-1}) \in \tau$ is the suffix of $\bz_{1}^{t-1}$.
\end{definition}

\subsection{Likelihood Function}

In order to extend the scope of VLMC models introduced previously to data involving multiple sequences,
we define a VLMC dataset of size $I$, denoted as $\cZ$, as a set of independent VLMC sequences
$\cZ = (\bZ^{(i)})_{i = 1, \dots, I}$ and $\cz = (\bz^{(i)})_{i = 1,\dots,I}$ will denote its observed realizations.

For each sequence $\bZ^{(i)}$, with length $T_i$, we will consider its first $L$ elements as constant values,
allowing us to write the joint probabilities as a product of transition probabilities
in \eqref{eq:cond_prob}, without requiring additional parameters for consistently defining the
probabilities of the first symbols in each sequence. Hence, the likelihood function is given by
\begin{equation}\label{eq:likelihood}
  f(\cz | \tau, \bp) =
  \prod_{\bs \in \tau} \prod_{k = 0}^{m-1} \left(p(k|\bs) \right)^{n_{\bs k}(\cz)},
\end{equation}
where $n_{\bs k}(\cz) = \sum_{i = 1}^I \sum_{t = L+1}^{T_i} \mathds{1}\left(z_t^{(i)} = k, \eta_\tau(\bz_{1}^{t-1(i)}) = \bs\right)$ counts the number of occurrences
of the symbol $k$ after strings with suffix $\bs$ across all sequences.

\subsection{Renewal States}

A symbol $a \in \cA$ is called a {\it renewal state} if
$$  \mathds{P}\left(\bZ_{t+1}^{t'} = \bz_{t+1}^{t'} | \bZ_1^{t-1} = \bz_{1}^{t-1}, Z_t = a \right) \,=\,   \mathds{P}\left(\bZ_{t+1}^{t'} = z_{t+1}^{t'} |  Z_t = a \right),$$
for all  $t > L$, $t' > t+1$, $\bz_{1}^{t-1} \in {\cal A}^{t-1}$, $\bz_{t+1}^{t'} \in {\cal A}^{t'-t}$.  That is, conditioning on $Z_t = a$, the distribution of the chain after $t$, $(Z_{u})_{u > t}$, is independent from
the past $(Z_u)_{u<t}$.

This property of conditional independences in a Markov Chain can be directly
associated to the structure of the context tree of a VLMC model.
For a VLMC model with associated context tree $\tau$, a state $a \in \cA$ is
a renewal state if  $a$ does not appear in any
inner node of the context tree. That is, for any context $\bs \in \tau$,
expressing $\bs$ as the concatenation of $l$ symbols $\bs = s_l\dots s_2 s_1$,
$s_{i} \neq a$ for $i = 1, \dots, l-1$.  In this case, we say that the tree $\tau$ is {\it $a$-renewing}.

Two out of the three trees displayed in \autoref{fig:renewal_examples} present renewal states. Tree (I) has $0$ as a renewal
state, Tree (II) has no renewal states due to the presence of the contexts $001$ and $00111$,
Tree (III) has only $1$ as a renewal state. If, in Tree (III),  the branch formed by contexts $002$, $102$, $202$ and $302$
was pruned and substituted by the context $02$ only, then $0$ would also be a renewal state.
Note that a VLMC may contain multiple renewal states.

A remarkable consequence of $a$ being a renewal state is that the random blocks between two occurrences of
$a$ are independent and identically distributed.
This feature allows the use of block Bootstrap methods,
enables a straightforward construction of cross-validation schemes and
 any other technique that relies on exchangeability properties.

\section{Bayesian Renewal Hypothesis Evaluation}

A VLMC model is fully specified by the probabilistic context tree $(\tau,\bp)$. The dimension of  $\bp$  depends on the branches of $\tau$.
Both these unobserved components $(\tau,\bp)$ can be treated as random elements with  given
prior distribution to carry out inference under the Bayesian paradigm.

From now on, we will use the following notation, for each $\bs \in \tau$,
$$\bp_{\bs} = (p(0|\bs), \ldots, p(m-1|\bs)) \in \Delta_m$$
where  $\Delta_m$ denotes the $m$-simplex,
\begin{equation*}
  \Delta_m = \left\{\mathbf{x} \in \mathds{R}^{\{0,1,\ldots,m-1\}} : \sum_{k = 0}^{m-1} x_k = 1 \text{ and } \forall j, x_j \geq 0\right\}.
\end{equation*}

In this section we discuss the prior specification for the probabilistic context tree $(\tau,\bp)$, as well as the resultant posterior distribution and how
to perform hypothesis testing using partial and intrinsic Bayes factor.

\subsection{A Bayesian Framework for VLMC models}

We consider a general prior distribution for $\tau$ proportional to any arbitrary non-zero function $h: \cT_L \rightarrow [0,\infty)$ and, given $\tau$,  for each $\bs \in \tau$, $\bp_{\bs}$ will have independent Dirichlet priors.

The complete Bayesian system can be described by the hierarchical structure
\begin{align}
  \tau &\sim \frac{h(\tau)}{\zeta(h, L)}, & \tau &\in \cT_L, \nonumber\\
  \bp | \tau &\sim \prod_{\bs \in \tau}\frac{\Gamma(\sum_{k = 0}^{m-1} \alpha_{\bs k})}{\prod_{k = 0}^{m-1}\Gamma(\alpha_{\bs k})}
  \prod_{k = 0}^{m-1} \left(p(k|\bs) \right)^{\alpha_{\bs k} - 1},
             & \bp &\in \Delta_m^{|\tau|}, \label{eq:hierarchical} \\
  \cZ | \tau, \bp &\sim f(\cz | \tau, \bp), & \bz^{(i)} &\in \cA^{T_i}, \nonumber
\end{align}
where
\begin{equation}
\label{eq:zeta}
\zeta(h,L) = \sum_{\tau \in \cT_L} h(\tau)
\end{equation}
 is the normalizing constant of the tree prior distribution
and $f(\cz|\tau, \bp)$ is given by \eqref{eq:likelihood}. We are assuming that the prior distribution for the transition probabilities $\bp_{\bs}$ are independent Dirichlet distribution with hyperparameters $\balpha_{\bs} = (\alpha_{\bs 1}, \ldots, \alpha_{\bs (m-1)})$.  Therefore, the joint distribution of $(\tau, \bp, \cZ)$ is given by
\begin{equation*}\label{eq:joint_distribution}
  \pi(\tau, \bp, \cz) =
  \frac{h(\tau)}{\zeta(h,L)}
    \prod_{\bs \in \tau} \frac{\Gamma(\sum_{k = 0}^{m-1} \alpha_{\bs k})}{\prod_{k = 0}^{m-1}\Gamma(\alpha_{\bs k})}
    \prod_{k = 0}^{m-1} \left(p(k|\bs)\right)^{n_{\bs k}(\cz) + \alpha_{\bs k} - 1}\,\mbox{.}
\end{equation*}

Since our interest lies in making inferences about the dependence structure represented by $\tau$ rather than the transition probabilities,
we can simplify our analysis by marginalizing the joint probability function over $\bp$,
$\pi(\tau, \cz) = \int \pi(\tau, \bp, \cz) d\bp$, obtaining a function that depends only on the context tree and the data.
The product of Dirichlet densities, assigned as the prior distribution of $\bp$, conjugates to the
likelihood function, allowing us to express the integrated distribution in closed-form as
\begin{equation}\label{eq:marginalized_distribution}
  \pi(\tau, \cz) = \frac{h(\tau)}{\zeta(h, L)} \prod_{\bs \in \tau}
  \frac{\Gamma(\sum_{k = 0}^{m-1} \alpha_{\bs k})}{\prod_{k = 0}^{m-1}\Gamma(\alpha_{\bs k})}
  \frac{\prod_{k = 0}^{m-1}\Gamma(n_{\bs k}(\cz) + \alpha_{\bs k})}{\Gamma(\sum_{k = 0}^{m-1} n_{\bs k}(\cz) + \alpha_{\bs k})},
\end{equation}
obtained by multiplying the appropriate normalizing constant to achieve Dirichlet densities with parameters
$(\alpha_{\bs0} + n_{\bs,0}(\cz), \dots, \alpha_{\bs (m-1)} + n_{\bs, m-1}(\cz))$
for each $\bs \in \tau$, so that the integration is done on a proper density.
For a less convoluted notation, we shall denote
\begin{equation}
\label{eq:q}
q(\tau, \cz) = \prod_{\bs \in \tau}
  \frac{\Gamma(\sum_{k = 0}^{m-1} \alpha_{\bs k})}{\prod_{k = 0}^{m-1}\Gamma(\alpha_{\bs k})}
  \frac{\prod_{k = 0}^{m-1}\Gamma(n_{\bs k}(\cz) + \alpha_{\bs k})}{\Gamma(\sum_{k = 0}^{m-1} n_{\bs k}(\cz) + \alpha_{\bs k})},
  \end{equation}
and use, from now on, the shorter expression
$\pi(\tau, \cz) = \frac{h(\tau)}{\zeta(h,L)} q(\tau, \cz)$.

Finally, the model evidence (marginal likelihood) can now be obtained by summing \eqref{eq:marginalized_distribution} over
all trees in $\cT_L$,
\begin{equation}\label{eq:evidence}
  \cE(\cz; h) = \sum_{\tau \in \cT_L} \pi(\tau, \cz) = \sum_{\tau \in \cT_L} \frac{h(\tau)}{\zeta(h, L)} q(\tau, \cz).
\end{equation}
Note that we explicitly describe the model evidence in terms of the prior distribution $h$ as we will be interested in
evaluating hypotheses based on different prior distributions.

\subsection{Bayes Factors for Renewal State Hypothesis}

Let $\cz = (\bz^{(i)})_{i = 1,\dots,I}$ be a VLMC sample compatible with  a probabilistic context tree $(\tau,\bp)$ where $\tau$ has maximum depth $L$. We will call {\it maximal tree} the complete tree with depth $L$ and let $a \in \cA$ be a fixed state of the alphabet. Our goal is to use Bayes Factors \citep{kass1995bayes} to evaluate the evidence in favor of the null hypothesis $H_a$
that $\tau$ is {\it $a$-renewing} against an alternative hypothesis $H_{\bar{a}}$ that $\tau$ is not {\it $a$-renewing}.
We denote  $\cT_{L}^{a} \subset \cT_L$ the set of $a$-renewing trees with depth no more than $L$
and $\bar\cT_L^a$ the set of trees with $a$ as an inner node and, consequently,  $a$ is not a renewal state for those trees.

We are interested in defining a metric for evaluating the hypothesis
$H_a:  \tau \in \cT_L^a$ against its complement
$H_{\bar a}: \tau \in \bar{\cT}_{L}^a$ in a Bayesian framework.
These hypotheses can be expressed in terms of special prior distributions proportional to functions
$h_a$ and $h_{\bar a}$, respectively, such that $h_a(\tau) = 0$ if, and only if,  $\tau \in \bar{\cT}_L^{a}$. Similarly,
$h_{\bar a}(\tau) = 0$ if, and only if,  $\tau \in \cT_L^{a}$.

The Bayes Factor for $H_a$ against $H_{\bar a}$ is defined as
\begin{equation}\label{eq:bayes_factor}
  \BF_{a,\bar{a}}(\cz) = \frac{\cE(\cz; h_a)}{\cE(\cz; h_{\bar a})} =
  \frac{\zeta(h_{\bar a}, L)}{\zeta(h_{a}, L)}
  \frac{
    \sum_{\tau \in \cT_L} h_a(\tau) q(\tau, \cz)
  }{
  \sum_{\tau \in \cT_L} h_{\bar a}(\tau) q(\tau, \cz)
},
\end{equation}
where $\zeta(\cdot,L)$ is given by \eqref{eq:zeta} and $q$ is given by \eqref{eq:q}.

\citet{kass1995bayes} proposed the following interpretation for the quantity $\log_{10}\left(BF_{a, \bar{a}}(\cz)\right)$ as a measure of evidence provided by the data $\cz$
in favor of the hypothesis that corresponds to $a$-renewing trees as opposed to the alternative one. A value between $0$ and $1/2$ is considered to provide evidence that is
``Not worth more than a bare mention", ``Substantial" for values between $1/2$ and $1$,
``Strong" if they are between $1$ and $2$ and ``Decisive" for values greater than $2$. By symmetry, these intervals with negative sign provide the same amount of evidence but reversing
the hypotheses considered. Therefore, the sign of $\log_{10}\left(BF_{a, \bar{a}}(\cz)\right)$
provides a straightforward measure whether the data provides more evidence that the chain is compatible with a context tree $\tau$ that is $a$-renewing
 or that $\tau$  belongs to $\bar{\cT}_L^{a}$.

\subsection{Metropolis-Hastings algorithm for context tree posterior sampling}

Before further development of methods to compute the Bayes Factors from \eqref{eq:bayes_factor},
we need to introduce a Metropolis-Hastings algorithm for sampling from the marginal posterior distribution of
context trees, $\pi(\tau|\cz)$. From \eqref{eq:marginalized_distribution} and the Bayes
rule we obtain
\begin{equation}
\label{eq:posterior}
  \pi(\tau|\cz) = \frac{\frac{h(\tau)}{\zeta(h, L)} q(\tau, \cz)}{\cE(\cz;h)} \propto h(\tau)q(\tau,\cz),
\end{equation}
which has a simple expression up to the intractable proportionality terms,
suggesting that the Metropolis-Hastings algorithm \citep{hastings1970monte, chib1995understanding}
is an appropriate strategy to obtain an empirical sample from the posterior distribution given by \eqref{eq:posterior}.

The main step for constructing the algorithm is defining a suitable proposal kernel
$\kappa(\tau'|\tau), \tau', \tau \in \cT_L$ to move to new context trees
from a current tree $\tau$. We propose the use of a graph-based kernel that can be
viewed as a modification of the
Monte Carlo Markov Chain Model Composition (MC$^3$) method from \citet{madigan1995bayesian}
by defining a neighborhood system $\mathcal{N}$ over $\cT_L$ and constructing a proposal kernel
that allows transitions only between neighboring trees only.

We first specify a set \textbf{directed} edges $\mathcal{N}_d$ such that, an edge $(\tau, \tau')$, from
$\tau$ to $\tau'$, is included if, and only if,
$|\tau \triangledown \tau'| = m+1$ and $|\tau'| > |\tau|$,
where $\triangledown$ denotes the symmetric difference operator
$A \triangledown B = (A \cap B^c) \cup (A^c \cap B)$.
An equivalent definition is that an edge from $\tau$ to $\tau'$ is obtained
substituting one of the contexts $\bs \in \tau$, by the contexts associated with its $m$
children nodes, $\{k\bs, k \in \cA\}$, in $\tau'$. We refer to this substitution as {\it growing
a branch}  from $\bs$.

Additionally, we define the  {\it grow} ($\grow$) and {\it prune} ($\ungrow$) operators, as
\begin{align*}
  \grow(\tau, h) &= \{ \tau' \in \cT_L: (\tau, \tau') \in \mathcal{N}_d \text{ and } h(\tau') > 0\}, \\
  \ungrow(\tau, h) &= \{ \tau' \in \cT_L: (\tau', \tau) \in \mathcal{N}_d \text{ and } h(\tau') > 0 \}.
\end{align*}
The operator  $\grow$ maps a tree $\tau$ to the set of trees with positive prior distribution
that can be obtained by growing new branches from $\tau$,
whereas $\ungrow$ maps $\tau$ to the set of trees in $\cT_L$ from which $\tau$ can be obtained after growing a branch. \\

Some important properties that can be easily checked are
\begin{enumerate}
  \item For every $\tau \in \cT_L$, if $\tau' \in \grow(\tau, h)$ and $h(\tau) > 0$, then $\tau \in \ungrow(\tau', h)$.
  \item For every $\tau \in \cT_L$, if $\tau' \in \ungrow(\tau, h)$ and $h(\tau) > 0$, then $\tau \in \grow(\tau', h)$.
  \item For any finite sequence $\tau^{(1)}, \tau^{(2)}, \dots, \tau^{(N)}$ such that $h(\tau^{(1)}) > 0$ and
    $\tau^{(n+1)} \in \grow(\tau^{(n)}, h) \cup \ungrow(\tau^{(n)}, h)$, we have
    $\tau^{(n)} \in \grow(\tau^{(n+1)}, h) \cup \ungrow(\tau^{(n+1)}, h)$.
\end{enumerate}
It follows from Properties 1 and 2  that any context tree $\tau$ can be recovered by applying sequentially
grow and prune operations.
Property 3 is a direct consequence of Properties 1 and 2 and means that any sequence of context trees
obtained by a sequence of grow or prune operations can also be visited in reverse
order with a sequence of grow and prune operations.
These properties also suggest that combining $\grow$ and $\ungrow$ for constructing a set of transitions with
positive probabilities in a proposal kernel is a good strategy in order to achieve the irreducibility condition
$\kappa(\tau|\tau') > 0$ if, and only if, $\kappa(\tau'|\tau) > 0$.

We define a transition kernel $\kappa$ as
\begin{equation*}
  \kappa(\tau'|\tau) =
  \begin{cases}
    \frac{1}{|\grow(\tau, h)|} \mathds{1}\left(\tau' \in \grow(\tau, h)\right), \text{ if }\ungrow(\tau, h) = \emptyset, \\
    \frac{1}{|\ungrow(\tau, h)|} \mathds{1}\left(\tau' \in \ungrow(\tau, h)\right), \text{ if }\grow(\tau, h) = \emptyset, \\
    \frac{1}{2} \frac{1}{|\grow(\tau, h)|} \mathds{1}\left(\tau' \in \grow(\tau, h)\right)  +
    \frac{1}{2} \frac{1}{|\ungrow(\tau, h)|} \mathds{1}\left(\tau' \in \ungrow(\tau, h)\right), \text{ o/w},\\
  \end{cases}
\end{equation*}
which allows us to propose a tree $\tau'$ in a simple two-step process. First, pick the operator to be applied to $\tau$,
$\grow$ or $\ungrow$ with probabilities $1/2$ if both lead to non-empty set of trees, otherwise pick the operation
that produces a non-empty set.
Then, pick $\tau'$ from $\grow(\tau, h)$ or $\ungrow(\tau,h)$
with uniform probabilities.

\begin{algorithm}[t]
  Set an initial tree $\tau^{(0)} \in \cT_L$\;
\For{$t = 1, \dots, \niter$}{
  Sample an operation $\grow$ or $\ungrow$ which leads to a non-empty set when applied to $\tau^{(t-1)}$,
  with equal probabilities\;
    Randomly pick a proposed tree $\tau'$ from $\grow(\tau^{(t-1)}, h)$ or $\ungrow(\tau^{(t-1)}, h)$\;
    Compute the acceptance ratio
    \begin{equation*}
      A(\tau'|\tau^{(t-1)}) =\min \left( \frac{h(\tau') q(\tau', \cz)}{h(\tau^{(t-1)}) q(\tau^{(t-1)}, \cz)}
      \frac{\kappa(\tau^{(t-1)}|\tau')}{\kappa(\tau'|\tau^{(t-1)})}, 1 \right);
    \end{equation*}
    Generate a random variable $U \sim \text{Unif}(0,1)$\;
    \eIf{$U < A(\tau|\tau^{(t-1)})$}{$\tau^{(t)} \leftarrow \tau' $}{$\tau^{(t)} \leftarrow \tau^{(t-1)}$}

}
\caption{Metropolis-Hastings algorithm for sampling context trees from $\pi(\tau|\cz)$ under a tree prior distribution proportional to $h$.}\label{algo:mh_tree}
\end{algorithm}

The idea of a proposal kernel for context trees based on growing and pruning nodes of trees was already used in \citet{kontoyiannis2020bayesian} for a specific prior distribution $h$.
The complete Metropolis-Hastings algorithm described in \autoref{algo:mh_tree}, not only formally defines the construction in terms of graphs, but also extends it to accommodate arbitrary prior distributions. 

\subsection{Partial Bayes Factors and Intrinsic Bayes Factors}

While $\BF_{a,\bar{a}}(\cz)$ given by \eqref{eq:bayes_factor} provides a measure of the plausibility of
one hypothesis with respect to another, computing this quantity may present an
enormous computational cost as the sum over $\cT_L$ involves a doubly exponential
number of terms. In fact, even the normalizing constant of the tree prior distribution
$\zeta(h,L)$ is intractable in the general case for moderate values $L$,
hindering the evaluation of the model evidence $\cE(\cz; h)$.

Previous algorithms proposed in the literature that use similar ideas to compute the marginal likelihood,
like the Context Tree Weighting (CTW) algorithm from \citet{willems1995context}, are not suitable for our purposes.
While we are aiming to compute \eqref{eq:evidence} for an arbitrary prior,
their weighting of context trees correspond to a very specific choice of prior distributions
as $h(\tau)$ such that \eqref{eq:evidence} can be computed recursively based on
the nodes of the maximal tree rather than every subtree.
To overcome this difficulty, we consider the Partial Bayes Factor (PBF) described in \citet{o1995fractional}
as an alternative approach for model comparison.

The methodology consists of dividing the data $\cz$ into two independent chunks,
$\cz_\train$ and $\cz_\test$ and then computing the Bayes Factor based on part of the data, $\cz_\test$,
conditioned on $\cz_\train$ as follows,
\begin{equation}\label{eq:partial_bayes_factors}
  \text{PBF}_{a, \bar{a}}(\cz_\test | \cz_\train) =
  \frac{
    \sum_{\tau \in \cT_L} \pi_a(\tau | \cz_\train) q(\tau, \cz_\test)
  }{
  \sum_{\tau \in \cT_L} \pi_{\bar{a}}(\tau | \cz_\train) q(\tau, \cz_\test)
},
\end{equation}
where $\pi_a(\tau | \cz_\train)$ and $\pi_{\bar{a}}(\tau | \cz_\train)$ are the posterior distributions of $\tau$ conditioned on the training data
$\cz_\train$ under the hypotheses $H_a$ and $H_{\bar{a}}$, respectively.

Although the original goal of using PBF is to avoid undefined behaviors when evaluating the ratio of terms involving improper priors, that are replaced by the posterior distributions conditioned on the training sample, we can see that the same strategy is very useful to avoid the intractable normalizing constant from the prior distribution.

Note that, even though \eqref{eq:partial_bayes_factors} still involves sums
over $\cT_L$, the terms
$$\sum_{\tau \in \cT_L} \pi_a(\tau | \cz_\train) q(\tau, \cz_\test) \quad \mbox{ and } \sum_{\tau \in \cT_L} \pi_{\bar{a}}(\tau | \cz_\train) q(\tau, \cz_\test) $$
can be written as expected values
$\mathds{E}_{H_a}\hspace{-1mm}\left(q(\tau, \cz_\test) | \cz_\train \right)$ and $\mathds{E}_{H_{\bar{a}}}\hspace{-1mm}\left(q(\tau, \cz_\test) | \cz_\train \right)$, which
can be obtained from ergodic Markov Chains $(\tau_a^{(t)})_{t \geq 1}$ and $(\tau_{\bar{a}}^{(t)})_{t \geq 1}$with invariant measures $\pi_a(\tau|\cz_\train)$ and $\pi_{\bar{a}}(\tau|\cz_\train)$, respectively.

Therefore, we can use MCMC methods to approximate Partial Bayes Factors by sampling two Markov Chains
$(\tau_a^{(t)})$ and $(\tau_{\bar{a}}^{(t)})$ and using the ratio
of empirical averages instead of the expected values
\begin{equation}\label{eq:pbf_approximation}
  \widehat{\text{PBF}}_{a, \bar{a}}(\cz_\test|\cz_\train) =
  \frac{
    \sum_{t = 1}^{\niter} q(\tau_a^{(t)}, \cz_\test)
  }{
  \sum_{t = 1}^{\niter} q(\tau_{\bar{a}}^{(t)}, \cz_\test)
}.
\end{equation}

To avoid the arbitrary segmentation of the dataset into train and test subsets,
\citet{berger1996intrinsic} proposed the Intrinsic Bayes Factor (IBF), which averages
PBFs obtained using different segmentations, based on minimal training samples.
Denote by $\mathcal{I}_v$ the collection of subsets of $\{1, \dots, I\}$ of size $v$,
i.e.,
\begin{equation*}
  \mathcal{I}_v = \{ \{i_1, i_2, \dots, i_v\} \subset \{1,2,\dots,I\} \}.
\end{equation*}
The dataset is divided into minimal training samples, which we will consider a $v$-tuple
of sequences $i_1^v = \{i_1, i_2, \dots, i_v\} \in \mathcal{I}_v$,
denoted $\cz^{(i_1^v)}$ and the remaining $I-v$ sequences,
denoted as $\cz^{(-i_1^v)}$ to be used as the test sample.
For each possible subset of $v$ sequences $i_1^v$,
we compute the Monte Carlo approximation of the PBF in \eqref{eq:pbf_approximation}
and take either the arithmetic average to obtain the Arithmetic Intrinsic Bayes Factor (AIBF)
or the geometric average for the Geometric Intrinsic Bayes Factor (GIBF).

Denoting by $(\tau^{(t)}_{a,i_1^v})$ and $(\tau^{(t)}_{\bar{a},i_1^v}), t = 1, \dots, \niter$  the Markov Chains of context trees obtained
using \autoref{algo:mh_tree} with target distribution $\pi_a(\tau|\bz^{(i_1^v)})$ and $\pi_{\bar{a}}(\tau|\bz^{(i_1^v)})$ (considering prior distribution proportional
to $h_a$ and $h_{\bar{a}}$), respectively, the AIBF  and GIBF are defined as
\begin{equation*}
  \AIBF_{a,\bar{a}}(\cz) = \sum_{i_1^v \in \mathcal{I}_v} \frac{1}{{{I}\choose{v}}}
  \left(
    \frac{\sum_{t = 1}^\niter q\left(\tau^{(t)}_{a, i_1^v}, \cz^{(-i_1^v)}\right)}{\sum_{t = 1}^\niter q\left(\tau^{(t)}_{\bar{a}, i_1^v}, \cz^{(-i_1^v)}\right)}
  \right),
\end{equation*}
and
\begin{equation*}
  \GIBF_{a,\bar{a}}(\cz) = \prod_{i_1^v \in \mathcal{I}_v}  \left(
      \frac{\sum_{t = 1}^\niter q\left(\tau^{(t)}_{a,i_1^v}, \cz^{(-i_1^v)}\right)}{\sum_{t = 1}^\niter q\left(\tau^{(t)}_{\bar{a}, i_1^v}, \cz^{(-i_1^v)}\right)}
    \right)^{{{I}\choose{v}}^{-1}}.
\end{equation*}
The complete procedure for obtaining these quantities for a given VLMC dataset is described in \autoref{algo:IBF}.

Note that, while a single sequence ($v = 1$) is theoretically sufficient to identify the context tree
and can be considered a minimal training sample, computing posterior distributions using small datasets
may result in posterior distributions that assign very low probabilities to context trees with long branches
due to smaller total counts on those longer branches.
Therefore, using more sequences (higher value for $v$), may lead to more consistent results as the posterior
distribution used in each PBF is more likely to capture long-range contexts.
On the other hand, the number of PBFs to be computed is $|\mathcal{I}_v| =$ ${I}\choose{v}$ which quickly
becomes prohibitive when $v$ increases.
The choice of $v$ is a trade-off between computational cost and the deepness of contexts to be captured by partial posterior distributions.

\begin{algorithm}[!htbp]
  \For{$i_1^v \in \mathcal{I}_v$}{
  Generate $\left(\tau^{(t)}_{a, i_1^v}\right), t = 1, \dots, \niter$ with the Context Tree Metropolis Hastings algorithm with target distribution $\pi_a(\tau|\bz^{(i_1^v)})$\;
  Generate $\left(\tau^{(t)}_{\bar{a}, i_1^v}\right), t = 1, \dots, \niter$ with target distribution $\pi_{\bar{a}}(\tau|\bz^{(i_1^v)})$\;
  Compute the Partial Bayes Factor for the $v$-tuple $i_1^v$
  \begin{equation*}
    \widehat{\text{PBF}}_{a, \bar{a}}(\cz^{(-i_1^v)}|\bz^{(i_1^v)}) =
    \frac{
      \sum_{t = 1}^{\niter} q(\tau_{a,i_1^v}^{(t)}, \cz^{(-i_1^v)})
    }{
    \sum_{t = 1}^{\niter} q(\tau_{\bar{a}, i_1^v}^{(t)}, \cz^{(-i_1^v)})
  }.
  \end{equation*}
}
Return the averages
\begin{align*}
  \text{AIBF}_{a,\bar{a}}(\cz) &= \frac{1}{{{I}\choose{v}}} \sum_{i_1^v \in \mathcal{I}_v} \widehat{\text{PBF}}_{a, \bar{a}}(\cz^{(-i_1^v)}|\bz^{(i_1^v)})\\
  \text{GIBF}_{a,\bar{a}}(\cz) &= \prod_{i_1^v \in \mathcal{I}_v} \left(\widehat{\text{PBF}}_{a, \bar{a}}(\cz^{(-i_1^v)}|\bz^{(i_1^v)})\right)^{{{I}\choose{v}}^{-1}}
\end{align*}
\caption{Complete algorithm for computing AIBF and GIBF with MCMC approximations of Partial Bayes Factors for a dataset $\cz$ based on $v$-tuples.}\label{algo:IBF}
\end{algorithm}

\section{Simulation Studies and Application}\label{sec:simul}

To show the strength of our method, we analyzed artificial VLMC
datasets generated from two binary models models and a real one coming from the field of Linguistics.

\subsection{Simulation for binary models}

In this section, the primary goal is to examine the performance of the AIBF and GIBF for evaluating the evidence in favor of a null hypothesis of $\tau$ being {\it $a$-renewing} considering aspects of the effect of the number of independent samples, the size of each chain, and discrimination ability when similar trees are considered.  We consider simulations for binary VLMC models with three different sample sizes $I = 3, 10, 25$. For each scenario we sample $I$ chains of equal
length, but three different values  $T_i = 1000, 2500, 5000$.
A dataset was simulated for each combination of $I$ and $T_i$, resulting in 9 datasets.

\begin{figure}[!htbp]
  \centering
  \begin{tikzpicture}[grow=down, scale=0.6]
  \tikzset{edge from parent/.style={draw,edge from parent path={(\tikzparentnode.south)-- +(0,-8pt)-| (\tikzchildnode)}},
    every leaf node/.style={font=\Large},
    every internal node/.style={draw,circle,font=\Large}}
    \Tree [ .{}
    [ .0 \edge[dashed]; .{$\left(\frac{1}{6}, \frac{5}{6}\right)$} ]
        [ .1
            [ .0 \edge[dashed]; {$\left(\frac{1}{2}, \frac{1}{2}\right)$} ]
            [ .1
              [ .0 \edge[dashed]; {$\left(\frac{1}{6}, \frac{5}{6}\right)$} ]
              [ .1
                [ .0 \edge[dashed]; {$\left(\frac{1}{2}, \frac{1}{2}\right)$} ]
                [ .1
                  [ .0 \edge[dashed]; {$\left(\frac{1}{6}, \frac{5}{6}\right)$} ]
                  [ .1
                    [ .0 \edge[dashed]; {$\left(\frac{1}{2}, \frac{1}{2}\right)$} ]
                    [ .1 \edge[dashed]; {$\left(\frac{1}{6}, \frac{5}{6}\right)$} ]]]]]]]
                    \node[above] at (current bounding box.north) {VLMC model 1};
  \end{tikzpicture}
  \begin{tikzpicture}[grow=down, scale=0.6]
  \tikzset{edge from parent/.style={draw,edge from parent path={(\tikzparentnode.south)-- +(0,-8pt)-| (\tikzchildnode)}},
    every leaf node/.style={font=\Large},
    every internal node/.style={draw,circle,font=\Large}}
    \Tree [ .{}
    [ .0 \edge[dashed]; {$\left(\frac{1}{6}, \frac{5}{6}\right)$} ]
        [ .1
            [ .0 \edge[dashed]; {$\left(\frac{1}{2}, \frac{1}{2}\right)$} ]
            [ .1
              [ .0 \edge[dashed]; {$\left(\frac{1}{6}, \frac{5}{6}\right)$} ]
              [ .1
                [ .0 \edge[dashed]; {$\left(\frac{1}{2}, \frac{1}{2}\right)$} ]
                [ .1
                  [ .0
                    [ .0 \edge[dashed]; {$\left(\frac{1}{6}, \frac{5}{6}\right)$} ]
                    [ .1 \edge[dashed]; {$\left(\frac{3}{4}, \frac{1}{4}\right)$} ]]
                  [ .1
                    [ .0 \edge[dashed]; {$\left(\frac{1}{2}, \frac{1}{2}\right)$} ]
                    [ .1 \edge[dashed]; {$\left(\frac{1}{6}, \frac{5}{6}\right)$} ]]]]]]]
                    \node[above] at (current bounding box.north) {VLMC model 2};
  \end{tikzpicture}
  \caption{Probabilistic Context Trees for Model 1 and Model 2. The pair of values below each leaf corresponds to the transition probabilities for the suffix associated with that leaf.}\label{fig:vlmc_models}
\end{figure}
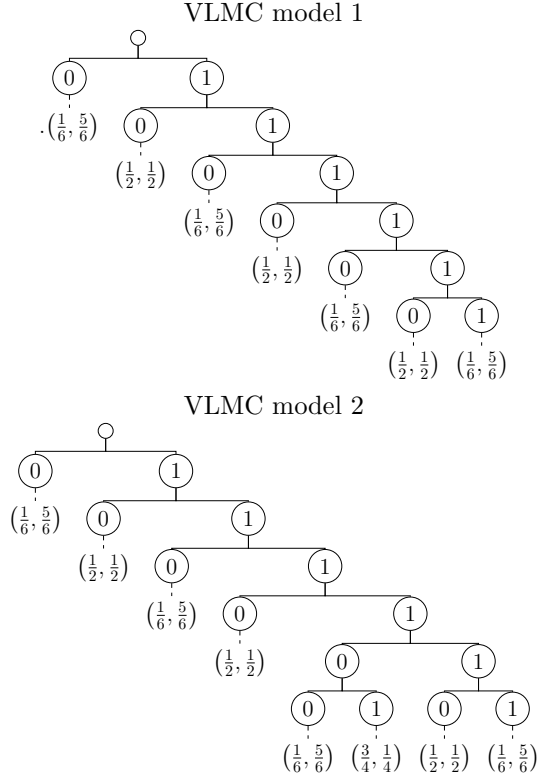

The two models considered are presented in \autoref{fig:vlmc_models}. Model 1 has
a depth equal to 6 with $a = 0$ being a renewal state, while Model 2 is a modified
version with an additional branch grown from the node $01111$, which is substituted
by the two suffixes $001111$ and $101111$. Therefore, in Model 2, $0$ is no longer
a renewal state although both trees are very similar.

\begin{table}[ht]
\centering
\caption{AIBF and GIBF computed in $\log_{10}$ scale for simulations for  Model 1.}\label{tab:sim_binary_1}
\begin{tabular}{c|cc|cc|cc|cc}
  \hline
  \multirow{2}{*}{$a$} & \multirow{2}{*}{$I$} & \multirow{2}{*}{$v$} & \multicolumn{2}{c}{$T_i = 1000$} & \multicolumn{2}{c}{$T_i = 2500$} & \multicolumn{2}{c}{$T_i = 5000$} \\
     & & & AIBF & GIBF & AIBF & GIBF & AIBF & GIBF \\
  \hline
  0 &   3 & 1 & 4.28 & 2.39 & 2.35 & 2.03 & 1.69 & 1.65 \\
  0 &   3 & 2 & 1.04 & 0.66 & 1.54 & 1.20 & 1.88 & 1.86 \\
  0 &  10 & 1 & 25.32 & 6.00 & 1.97 & 1.94 & 2.55 & 2.53 \\
  0 &  10 & 2 & 4.05 & 2.31 & 1.96 & 1.90 & 2.49 & 2.46 \\
  0 &  25 & 1 & 68.98 & 11.87 & 4.40 & 2.90 & 2.59 & 2.57 \\
  0 &  25 & 2 & 67.29 & 3.54 & 2.63 & 2.60 & 2.59 & 2.57 \\
  \hline
  1 &   3 & 1 & -46.87 & -49.03 & -137.09 & -147.48 & -285.28 & -287.18 \\
  1 &   3 & 2 & -17.82 & -20.41 & -61.80 & -68.78 & -135.72 & -138.05 \\
  1 &  10 & 1 & -248.24 & -265.45 & -683.99 & -695.28 & -1401.88 & -1411.24 \\
  1 &  10 & 2 & -234.44 & -238.54 & -596.78 & -616.72 & -1237.09 & -1252.95 \\
  1 &  25 & 1 & -688.51 & -741.06 & -1921.76 & -1951.27 & -3834.79 & -3862.61 \\
  1 &  25 & 2 & -705.66 & -719.25 & -1822.97 & -1869.77 & -3653.59 & -3701.06 \\   \hline
\end{tabular}
\end{table}

\begin{table}[ht]
\centering
\caption{AIBF and GIBF computed in $\log_{10}$ scale for simulations for Model 2.}\label{tab:sim_binary_2}
\begin{tabular}{c|cc|cc|cc|cc}
  \hline
  \multirow{2}{*}{$a$} & \multirow{2}{*}{$I$} & \multirow{2}{*}{v} & \multicolumn{2}{c}{$T_i = 1000$} & \multicolumn{2}{c}{$T_i = 2500$} & \multicolumn{2}{c}{$T_i = 5000$} \\
     & & & AIBF & GIBF & AIBF & GIBF & AIBF & GIBF \\
  \hline
0 &   3 & 1 & 2.58 & 2.45 & 14.38 & 0.06 & -7.51 & -8.24 \\
  0 &   3 & 2 & 1.34 & 1.22 & 7.40 & 0.05 & -2.94 & -3.29 \\
  0 &  10 & 1 & 13.58 & 4.13 & 26.40 & -7.77 & 70.50 & -38.47 \\
  0 &  10 & 2 & 12.49 & 2.39 & 26.27 & -2.94 & 65.34 & -34.02 \\
  0 &  25 & 1 & 46.19 & 9.11 & 94.14 & 13.73 & 230.50 & -95.80 \\
  0 &  25 & 2 & 46.49 & 5.54 & 93.36 & -26.54 & 227.65 & -111.42 \\ \hline
  1 &   3 & 1 & -38.24 & -40.92 & -101.68 & -114.08 & -216.82 & -233.08 \\
  1 &   3 & 2 & -15.16 & -18.32 & -52.24 & -54.20 & -111.01 & -113.50 \\
  1 &  10 & 1 & -215.74 & -222.35 & -539.57 & -573.34 & -1062.60 & -1167.87 \\
  1 &  10 & 2 & -190.76 & -201.17 & -479.56 & -509.43 & -930.81 & -1041.44 \\
  1 &  25 & 1 & -566.23 & -601.98 & -1432.13 & -1528.50 & -2867.44 & -3186.70 \\
  1 &  25 & 2 & -535.09 & -583.43 & -1365.51 & -1502.80 & -2746.98 & -3091.29 \\
\hline
\end{tabular}
\end{table}

\begin{figure}[h]
  \centering
  \includegraphics[width=0.8\linewidth]{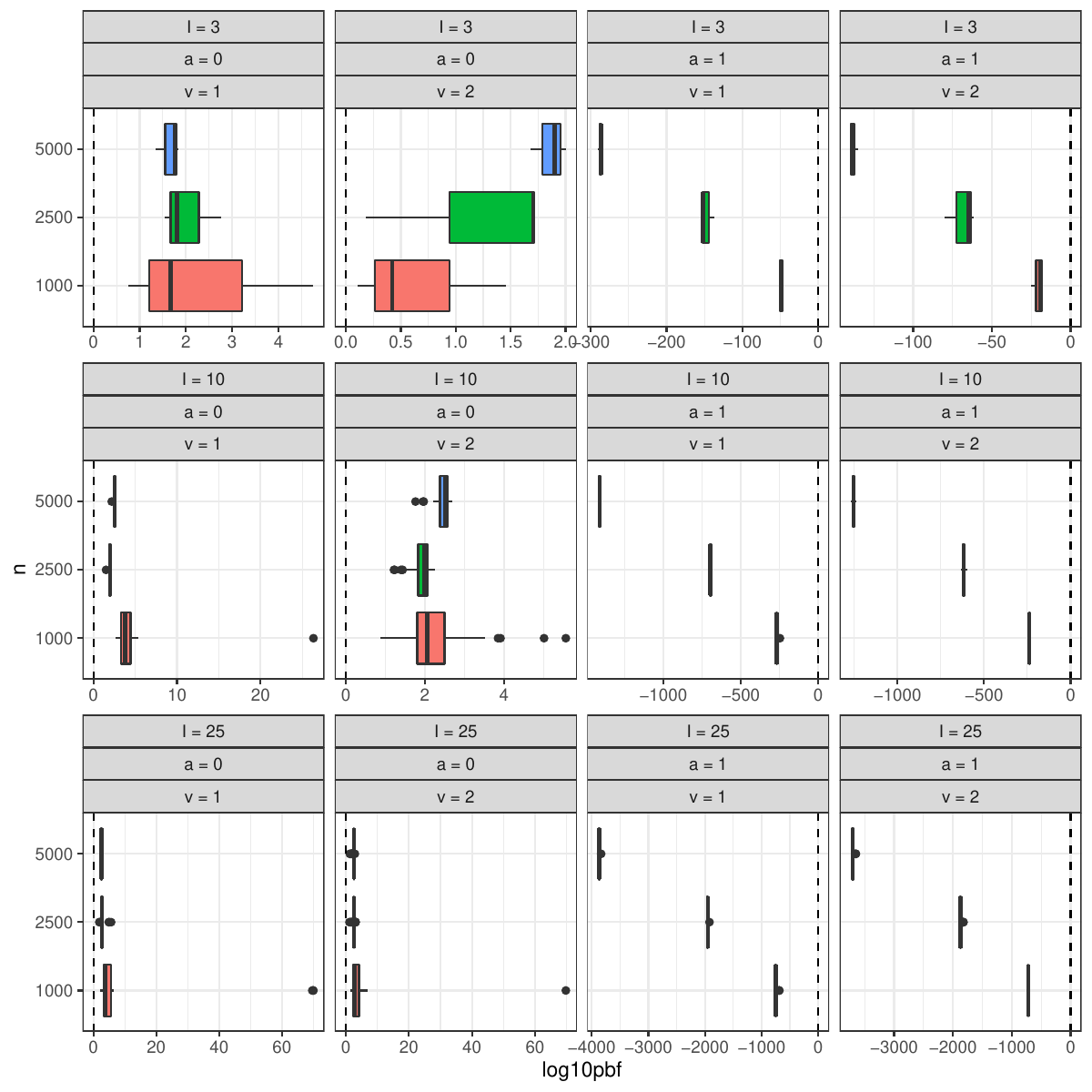}
  \caption{Computed PBFs for Model 1 of the simulation study.
  Only $0$ is a renewal state for this model, therefore, the panels with $a = 0$
  are expected to have positive PBFs in logarithmic scale and negative values for $a = 1$.}
  \label{fig:6}
\end{figure}

\begin{figure}[h]
  \centering
  \includegraphics[width=0.8\linewidth]{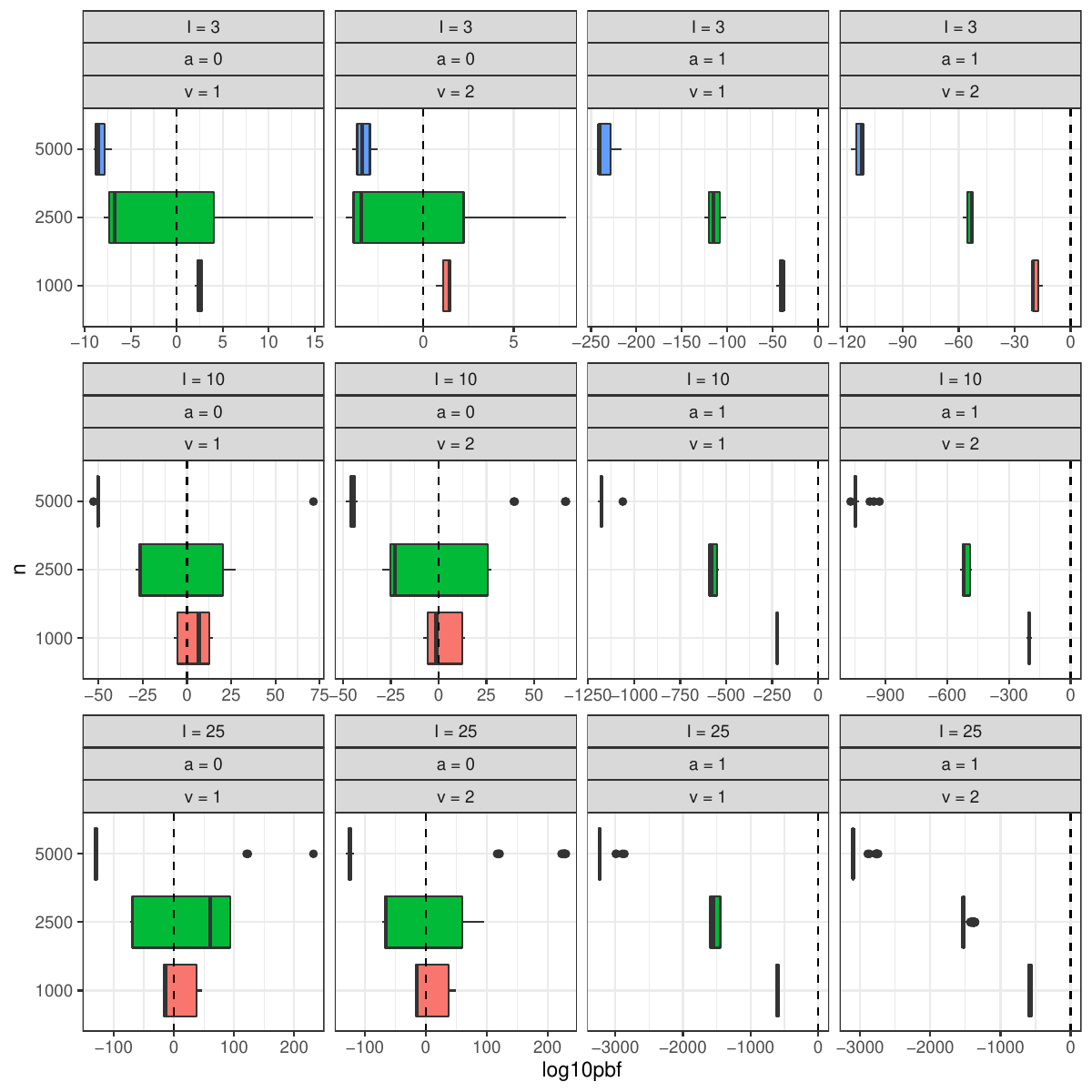}
  \caption{Computed PBFs for Model 2 of the simulation study.
  Neither $0$ or $1$ are renewal states for this model, therefore, all panels
  are expected to have negative PBFs in logarithmic scale.
  The structure of the tree used makes the renewal state hypothesis violation for
  $a = 0$ to not be captured (or does not produce strong evidence against the renewal state hypothesis)
  for lower sample sizes.} \label{fig:7}
\end{figure}

For each possible renewal state $a = 0$ or $a = 1$,
we compute both Intrinsic Bayes Factors (AIBF and GIBF) using \autoref{algo:IBF} considering  prior distributions
proportional to
\begin{equation*}\label{eq:prior_a}
  h_a(\tau) = \mathds{1}(\tau \in \cT_L^a)
  \hspace{1cm}\text{ and }\hspace{1cm}
  h_{\bar a}(\tau) = \mathds{1}(\tau \in \bar{\cT}_L^{a})
\end{equation*}
which correspond to the uniform distribution in the space of context trees that
are allowed under $H_a$ and $H_{\bar a}$, respectively.

For the  hyper-parameter $\balpha$, we choose $\alpha_{\bs k} = 0.001$ for all $\bs$ and $k$,
resulting in symmetrical prior distributions for the transition probabilities and a higher density for vectors that
are more concentrated.

In each scenario, we considered $v = 1$ and $v = 2$ as the number of sequences to be used for the minimal training sample and ran
$\niter = 10^5$ Metropolis-Hastings steps for each PBF Monte Carlo approximation.

From \autoref{tab:sim_binary_1}, we can see that, for Model 1, both AIBF and GIBF lead to the correct decision for both renewal states tested.
For $a = 0$, which was in fact a renewal state, all cases returned a value, in $\log_{10}$ scale,
greater than 1 (except one equals $0.66$)
The discrepancy between AIBF and GIBF diminishes as $T_i$ and $I$ increase converging to a value around 2.5  which was considered Decisive Evidence in the Kass-Raftery scale.
On the other hand, for $a=1$, which was not a renewal state, all cases reported AIBF and GIBF in $\log_{10}$ scale smaller than $-15$, converging to values around $-3000$ when $T_i=500$ and $I=25$.

The results for Model 2, which includes a new pair of contexts causing 0 to be no longer a renewal state, presented in \autoref{tab:sim_binary_2},
showed a similar performance rejecting the 1-renewing hypothesis when compared to Model 1,
which is expected due to the similarity of both models with respect to the short-length nodes that do
not involve $1$. The main difference occurred in the computed values for the 0-renewing hypothesis, where the computed
GIBFs were positive in logarithmic scale for $T_i = 1000$, for all values of $I$ and $v$.
Moreover, we can see that for $T_i=5000$, GIBF was negative for all scenarios whereas AIBF was strongly positive for larger values of $I$.
For intermediate value of the sequences sizes ($T_i=2500$), AIBF pointed to the wrong direction
in all scenarios while GIBF identified the right hypothesis in three cases ($I=10$ and $v=1$ and 2 and $I=25$ and $v=2$).
This suggests that for smaller sample sizes, the posterior distributions obtained from the training
samples were insufficient to capture the long-range contexts $111100$ and $111101$ that break the renewal condition of the state 0.

Figures \ref{fig:6} and \ref{fig:7} present the empirical distributions of PBFs computed in $\log_{10}$ scale for each scenario and each model.
In general, the strength of the evidence tends to be larger for $v = 1$ as more data is being used on the test set,
but on the other hand, $v = 2$ leads to more stable PBFs computed. The same behavior is observed  as
the number of independent sequences $I$ increases, what is expected as adding more data has an impact
in the scale of the marginal likelihood function, also rescaling the Bayes Factors.

With $T_i = 2500$, $\log_{10}$ PBFs tend to be distributed around $0$, with high variance, as can be observed in \autoref{fig:7},
resulting in a very unstable average, leading to correct results in some of the scenarios, and incorrect ones in others.
As the sample size increases, those long-range contexts are more likely to be captured by the posterior distribution.
With $T_i = 5000$, we have decisive evidence that $0$ is not a renewal state, except for the scenario with $I = 3$ and $v = 1$,
where the value of $0.13$ provides very weak evidence, and, in general, the distribution of $\log_{10}$ PBFs was highly concentrated with negative sign,
except for a few outliers. AIBF was highly affected by outliers and resulted in incorrect conclusions with high evidence for some scenarios.

Note that, especially for the datasets with small samples ($T_i = 1000$), outliers with large values are observed, having
great effect on the computed averages, although the conclusions are not affected. For large datasets ($T_i = 5000$),
we have smaller variance in the computed PBFs among different training sets compared to scenarios with sequences of smaller sizes.

Therefore, we conclude that a decision based on GIBF leads to, at least, strong evidence in the scale from \citet{kass1995bayes}
(greater than $1/2$ in $\log_{10}$ scale)
for the correct hypothesis in all cases where we had $v = 2$ and $T_i = 2500$ or $5000$.
The AIBF was not robust to the presence of outliers in the set of PBFs,
leading to incorrect conclusions in the scenarios where
the correct detection of the renewal state is harder task and the sample sequences are shorter.
The present of outliers also suggests other functions to summarize PBFs other than geometric and arithmetic
averages may be useful for avoiding having results highly influenced for the results obtained for particular
test samples, like trimmed averages, removing the most extreme values from the IBF computation, or using
the median PBF, which corresponds to a trimmed average trimming all but one value, as used in \citet{charitidou2018objective}.
Context trees that break renewal state condition on long-ranged contexts require larger samples ($T_i$ or $v$) in order to
be captured and result in evidence against the renewal state hypothesis,
while states that break renewal state condition in short contexts can be immediately identified,
even with short observed sequences.

\subsection{Application to rhythm analysis in Portuguese texts}

It is known that Brazilian and European Portuguese (henceforth BP and EP) have different syntaxes. For example, \cite{galves2005syntax} infered that the placement of clitic pronouns in BP and EP differ in two senses, one of them being: ``EP clitics, but not BP clitics, are required to be in a non-initial position with respect to some boundary".  However, the question remains: are the choices of word placement  related to  different stress patterns preferences?   This question was addressed by \citet{galves2012context} that found distinguishing rhythmic patterns for BP and EP based on written journalistic texts. The data consists of 40 BP texts and 40 EP texts randomly extracted from an encoded corpus of newspaper articles from the 1994 and 1995 editions of {\it Folha de S\~ ao Paulo} (Brazil) and {\it O P\'ublico} (Portugal). Texts were encoded according on rhythmic features resulting in
discrete sequences with around 2500 symbols each and are available at \url{http://dx.doi.org/10.1214/11-AOAS511SUPP}. After a preprocessing of the texts (removing foreign words, rewriting of symbols, dates, compound words, etc) the syllables were  encoded  by assigning one of four symbols according to whether or not
(i) the syllable is stressed;
(ii) the syllable is the beginning of a prosodic word (a lexical word (noun, verbs,...) together with the functional
non-stressed words (articles, prepositions, ...) which precede or succeed it). This classification can be represented by  4 symbols.  Additionally  an extra symbol was assigned to encode the end of each sentence.
The alphabet $\mathcal{A} = \{0, 1, 2, 3, 4\}$ was obtained as follows.
\begin{itemize}
\item 0 = non-stressed, non-prosodic word initial syllable;
\item 1 = stressed, non-prosodic word initial syllable;
\item 2 = non-stressed, prosodic word initial syllable;
\item 3 = stressed, prosodic word initial syllable;
\item 4 = end of each sentence.
\end{itemize}

 For example, the sentence {\it O sol brilha forte agora.} (The sun shines bright now.) is coded as

\begin{tabular}{lcccccccccc}
Sentence   & {O} & {sol} & {bri} & {lha} & {for} & {te} & {a} &  {go} & {ra} & .\\
Code       & 2 & 1  & 3  & 0  & 3      & 0  &  2  & 1  & 0  & 4\\
\end{tabular}

\vspace{2mm}
The Smallest Maximizer Criteria proposed by \citet{galves2012context} to select the best tree for BP and EP,  uses the fact that the symbol $4$ appears as a renewal state to perform Bootstrap sampling. Moreover, they conclude that ``the main difference between the two languages is that whereas in BP both 2 (unstressed boundary of a phonological
word) and 3 (stressed boundary of a phonological word) are contexts, in EP only 3 is a context."  These are exactly the type of questions to be addresses by the renewal state detection algorithm.

Due to the encoding used, the grammar of the language, and the general structure of written texts, some transitions are
not possible. For example, two end of sentences (symbol 4) cannot happen consecutively, therefore,
a transition from 4 to 4 is not allowed.
Furthermore, there is one, and only one, stressed syllable in each prosodic word.
 \autoref{tab:allowed} summarizes the allowed and prohibited one-step transitions.

\begin{table}[!htbp]
  \centering
  \caption{Allowed transitions for each encoded symbol.}\label{tab:allowed}
  \begin{tabular}{c|ccccc}
    From/To & 0 & 1 & 2 & 3 & 4 \\ \hline
    0 & yes & yes & yes & yes & yes \\
    1 & yes & no & yes & yes & yes \\
    2 & yes & yes & no & no & no \\
    3 & yes & no & yes & yes & yes \\
    4 & no & no & yes & yes & no \\
  \end{tabular}
\end{table}

These prohibited transition conditions are included in the model with proper modifications
to the prior distribution, assigning zero probability to some context trees and
forcing the probabilities related to prohibited transitions to be zero.
The modifications are:
\begin{enumerate}
  \item If a transition from $k$ to $k'$ is prohibited and a context $\bs$ has $k$ as its last symbol,
    we force $p_{\bs,k} = 0$  in our prior distribution.
    The remaining probabilities associated with allowed transitions are then a priori distributed as a Dirichlet distribution
    with lower dimension.

    For example, for a context $102$, the only allowed transitions from $2$ are to $0$ and $1$, we have
    $p_{102, 2} = p_{102, 3} = p_{102, 4} = 0$ with prior probability $1$,
    and the free probabilities, $(p_{102, 0}, p_{102, 1})$, distributed as a 2-dimensional Dirichlet distribution with
    hyper-parameters $(\alpha_{102,0}, \alpha_{102,1})$.

  \item If $\bs \in \tau$ includes a prohibited transition, then $n_{\bs,k}(\cz) = 0$ as there will not be any occurrences
    of such sequence in the sample. As a consequence, these suffixes have no contribution in $q(\tau, \cz)$ as the term related to $\bs$ of the product in \eqref{eq:marginalized_distribution} gets cancelled.

  \item We define $\cT^*_{L}$ the space of context trees that do not have prohibited transitions
in inner nodes. For example, since the transition $44$ is prohibited, a tree that contains a suffix $044$
cannot be in $\cT^*_{L}$ because the transition from $4$ to $4$ is not in a leaf (final node),
whereas a tree in $\cT_{L}$ can contain the suffix $44$.

Note that, allowing final nodes to contain a prohibited transition is necessary to keep the
consistency of our definition based on \textbf{full} $m$-ary trees, as full tree contains
the suffix $34$ (allowed) if, and only if, it contains another suffix ending in
$44$ (prohibited), what is not a problem because this prohibited suffix will not contribute
to the marginal likelihood.
\end{enumerate}

For each set of sequences in BP and EP,
we compute Intrinsic Bayes Factors as evidence for the five hypotheses that
0, 1, 2, 3, and 4 are renewal states.
We took $L = 5$ which should be enough to cover all
relevant context trees based on the results from the original paper.
For prior distributions, we
used the same uniform distributions as in the simulation experiment, but restricted to the trees that do not include prohibited transitions on inner nodes, i.e.,
\begin{equation*}\label{eq:prior_pt}
  h_a(\tau) = \mathds{1}(\tau \in \cT_5^{a} \cap \cT^*_5)
  \hspace{1cm}\text{ and }\hspace{1cm}
  h_{\bar a}(\tau) = \mathds{1}(\tau \in \bar{\cT}_5^{a} \cap \cT^*_5).
\end{equation*}
We also set $\alpha_{\bs k} = 0.001$ for every $\bs \in \tau$ and $k \in \cA$.
The algorithm ran for $\niter = 10^6$ iterations for computing each tree posterior
distribution under each hypothesis, using $v=2$ sequences
(around 5000 symbols in each training sample) for
each Partial Bayes Factor, resulting in a total of
${40}\choose{2}$ $= 780$ posterior distributions for each hypothesis and
PBFs to average.

\begin{figure}[!htbp]
  \centering
  \includegraphics[width=0.8\linewidth]{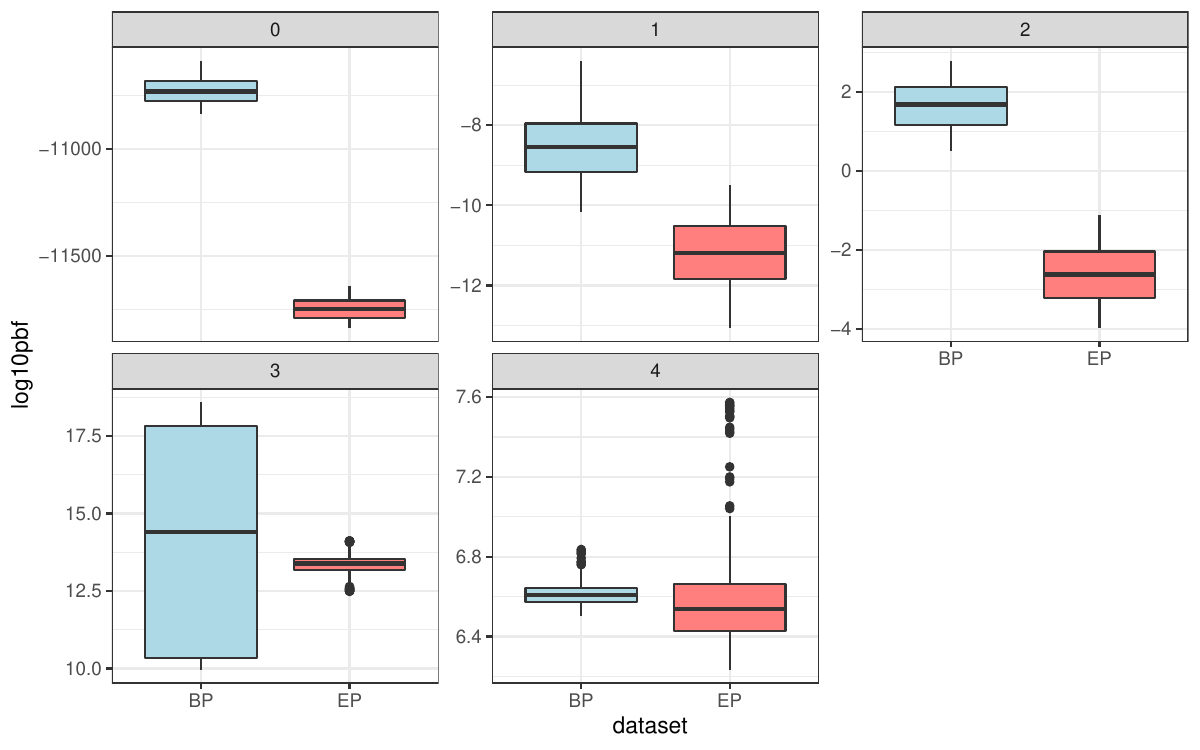}
  \caption{Boxplots presenting the distribution of estimated Partial Bayes Factors $\widehat{\text{PBF}}$ for the ${40}\choose{2}$ pairs of sequences used as training sets, in each of the BP and EP datasets after a trimming of 10\% of the highest and lowest PBFs. Each pane corresponds to a hypothesis of a different state being evaluated as a renewal state.} \label{fig:pbf_pt}
\end{figure}

Due to the numerical instabilities caused by outliers in the set of estimated PBFs as identified in the
simulation study, especially in the AIBF, when some particular texts
$\bz^{(i)}$ are used as the training sample, we also computed a
trimmed version of AIBF (and GIBF), which consists of computing
the arithmetic (and geometric) average excluding the $10\%$ lowest and
$10\%$ highest PBFs, this strategy was also used in \citet{berger1996intrinsic}.
The empirical distributions for the estimated Partial Bayes Factors for all renewal state hypotheses
after the$10\%$ trimming are shown in \autoref{fig:pbf_pt}.

From the results presented in \autoref{tab:portuguese},
we can see that decisive evidence was obtained when evaluating the renewal hypothesis
for states 2, 3 and 4 for the BP dataset and 3 and 4 for the EP dataset which are consistent with the results from \citet{galves2012context}.

\begin{table}[!htbp]
\centering
\caption{AIBF and GIBF for the BP and EP datasets.}\label{tab:portuguese}
\begin{tabular}{c|c|rr|rr}
  \hline
   &  & \multicolumn{2}{c|}{$\text{AIBF}$}  & \multicolumn{2}{c}{$\text{GIBF}$} \\
  \hline
  & $a$ & Untrimmed & Trimmed & Untrimmed & Trimmed \\ \hline
BP & 0 & -10321.12 & -10591.73 & -10720.48 & -10727.80 \\
  BP & 1 & 5.88 & -7.92 & -7.56 & -8.55 \\
  BP & 2 & 3.49 & 2.00 & 1.64 & 1.64 \\
  BP & 3 & 19.41 & 17.69 & 14.17 & 14.10 \\
  BP & 4 & 7.96 & 6.61 & 6.68 & 6.61 \\ \hline
  EP & 0 & -11487.31 & -11641.82 & -11744.90 & -11747.73 \\
  EP & 1 & 7.87 & -10.51 & -11.67 & -11.19 \\
  EP & 2 & 1.52 & -2.05 & -2.56 & -2.60 \\
  EP & 3 & 18.37 & 13.49 & 13.40 & 13.37 \\
  EP & 4 & 12.52 & 6.66 & 6.64 & 6.57 \\
   \hline
\end{tabular}
\end{table}

\begin{figure}[!htbp]
  \centering
  \begin{tikzpicture}[grow=down, scale=0.66]
  \tikzset{edge from parent/.style={draw,edge from parent path={(\tikzparentnode.south)-- +(0,-8pt)-| (\tikzchildnode)}},
    every internal node/.style={draw,circle}}
    \Tree [ .{}
      [ .0
          [ .0
            [ .0
              [ .0 ]
              [ .2 ]
            ]
            [ .1 ]
            [ .2 ]
            [ .3 ]
          ]
          [ .1
            [ .0 ]
            [ .2 ]
          ]
          [ .2 ]
          [ .3 ]
      ]
      [ .1
        [ .0
          [ .0 ]
          [ .2 ]
        ]
        [ .2 ]
      ]
      [ .2
        [ .0 ]
        [ .1 ]
        [ .3 ]
        [ .4 ]
      ]
      [ .3 ]
      [ .4 ]]
                  \node[above] at (current bounding box.north) {$\pi(\tau | \bz) = 0.776$};
  \end{tikzpicture}
  \begin{tikzpicture}[grow=down, scale=0.66]
    \tikzset{edge from parent/.style={draw,edge from parent path={(\tikzparentnode.south)-- +(0,-8pt)-| (\tikzchildnode)}},
      every internal node/.style={draw,circle}}
      \Tree [ .{}
      [ .0
          [ .0
            [ .0 ]
            [ .1 ]
            [ .2 ]
            [ .3 ]
          ]
          [ .1
            [ .0 ]
            [ .2 ]
          ]
          [ .2 ]
          [ .3 ]
      ]
      [ .1
        [ .0
          [ .0 ]
          [ .2 ]
        ]
        [ .2 ]
      ]
      [ .2
        [ .0 ]
        [ .1 ]
        [ .3 ]
        [ .4 ]
      ]
      [ .3 ]
      [ .4 ]]
                    \node[above] at (current bounding box.north) {$\pi(\tau | \bz) = 0.083$};
    \end{tikzpicture}
  \caption{Highest posterior probability context trees for the EP dataset.}\label{fig:ep_posterior}
\end{figure}
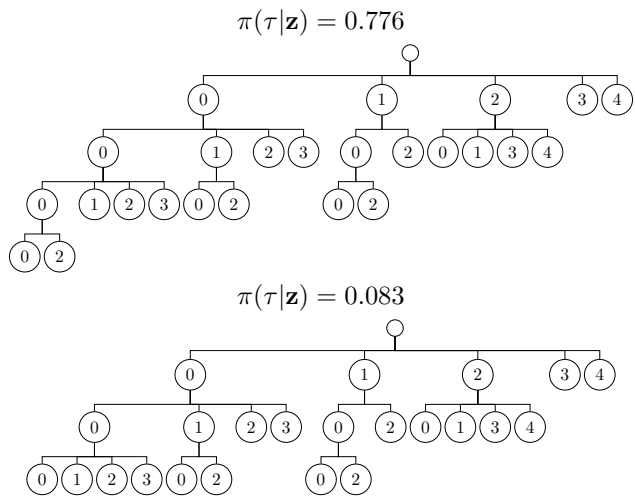

Finally, for completeness of the Bayesian analysis of this example, we ran the Metropolis Hastings algorithm for the entire BP and EP datasets, considering the same uniform
prior distribution on every tree in $\cT^*_5$
(no renewal state hypothesis considered, but controlling for prohibited transitions)
and with $\alpha_{\bs,k} = 0.001$ for each valid pair $(\bs,k)$.
The two context trees with highest posterior probabilities for BP and EP are presented in
\autoref{fig:bp_posterior} and \autoref{fig:ep_posterior}, respectively.
Leaves corresponding to contexts that contain prohibited transitions,
or other contexts with no occurrences are omitted from the trees in the figures
for interpretability purposes.

\begin{figure}[!htbp]
  \centering
  \begin{tikzpicture}[grow=down, scale=0.66]
  \tikzset{edge from parent/.style={draw,edge from parent path={(\tikzparentnode.south)-- +(0,-8pt)-| (\tikzchildnode)}},
    every internal node/.style={draw,circle}}
    \Tree [ .{}
      [ .0
          [ .0
            [ .0
              [ .0 ]
              [ .2 ]
            ]
            [ .1 ]
            [ .2 ]
            [ .3 ]
          ]
          [ .1
            [ .0
              [ .0 ]
              [ .2 ]
            ]
            [ .2 ]
          ]
          [ .2 ]
          [ .3 ]
      ]
      [ .1
        [ .0
          [ .0 ]
          [ .2 ]
        ]
        [ .2 ]
      ]
      [ .2 ]
      [ .3 ]
      [ .4 ]]
                  \node[above] at (current bounding box.north) {$\pi(\tau | \bz) = 0.500$};
  \end{tikzpicture}
  \begin{tikzpicture}[grow=down, scale=0.66]
    \tikzset{edge from parent/.style={draw,edge from parent path={(\tikzparentnode.south)-- +(0,-8pt)-| (\tikzchildnode)}},
      every internal node/.style={draw,circle}}
      \Tree [ .{}
        [ .0
            [ .0
              [ .0 ]
              [ .1 ]
              [ .2 ]
              [ .3 ]
            ]
            [ .1
              [ .0
                [ .0 ]
                [ .2 ]
              ]
              [ .2 ]
            ]
            [ .2 ]
            [ .3 ]
        ]
        [ .1
          [ .0
            [ .0 ]
            [ .2 ]
          ]
          [ .2 ]
        ]
        [ .2 ]
        [ .3 ]
        [ .4 ]]
                    \node[above] at (current bounding box.north) {$\pi(\tau | \bz) = 0.356$};
    \end{tikzpicture}
  \caption{Highest posterior probability context trees for the BP dataset.}\label{fig:bp_posterior}
\end{figure}
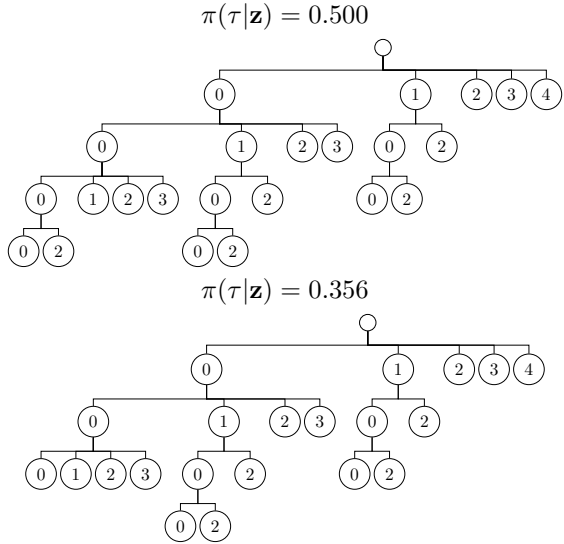

\section{Conclusions}

We propose a Bayesian approach to Variable-Length Markov Chain models with random context trees that
allows us to evaluate evidence in favor of renewal hypothesis based on a priori distributions that assign positive probability to each tree on
a subset of trees that depends on the renewal state being considered. The main novelties of this work are:
\begin{enumerate}
\item The use of Bayes Factor to test the renewal hypothesis for VLMC models;
\item The use of Intrinsic Bayes Factor to evaluate this evidence and overcame the problem of intractable normalizing constant from the prior distribution;
\item The proposal of a  Metropolis-Hastings algorithm for sampling context trees that can be performed under a tree prior distribution proportional to any arbitrary function $h$.
This freedom allows not only to incorporate experts prior information about the possible context trees, but also to exclude forbidden trees just by assigning zero probability to them.
\end{enumerate}

To show the strength of our method, we analyzed artificial
datasets generated from two binary models and one example coming from the field of Linguistics. The analysis of the artificial datasets suggests that evidence becomes stronger as the number of replicates increases and/or the size of the sequences increases. However, it is possible to obtain good results with as few as 3 replicates of each chain.  In the linguistics example we could observe that trimmed GIBF is more robust to  possible outliers in the sample.

An R package containing functions for all the computations used in this work is available at \href{github.com/Freguglia/ibfvlmc}{github.com/Freguglia/ibfvlmc} and R scripts used to reproduce the simulation study are available upon request.

\section*{Acknowledgments}

This work was funded by Funda\c c\~ao de Amparo \`a Pesquisa do Estado de S\~ao
Paulo - FAPESP grants 2017/25469-2 and 2017/10555-0, CNPq grant 304148/2020-2 and by Coordena\c c\~ao de Aperfei\c coamento
de Pessoal de N\'\i vel Superior - Brasil (CAPES) - Finance Code 001. We  thank Helio Migon and Alexandra Schmidt for fruitful discussions.

\bibliography{main.bib}
\bibliographystyle{apalike}
\pagebreak

\end{document}